\documentclass{article}

    \usepackage[final, nonatbib]{tackling_climate_workshop}

\usepackage[utf8]{inputenc} % allow utf-8 input
\usepackage[T1]{fontenc}    % use 8-bit T1 fonts
\usepackage{hyperref}       % hyperlinks
\usepackage{url}            % simple URL typesetting
\usepackage{booktabs}       % professional-quality tables
\usepackage{amsfonts}       % blackboard math symbols
\usepackage{nicefrac}       % compact symbols for 1/2, etc.
\usepackage{microtype}      % microtypography

\usepackage{graphicx} % ILIAS ADD
\usepackage{floatrow} %ilias
\newfloatcommand{capbtabbox}{table}[][\FBwidth] %ilias

\title{Causality and Explainability for Trustworthy Integrated Pest Management}

\author{%
Ilias Tsoumas\thanks{Equal contribution.} $^{1,2}$ \quad Vasileios Sitokonstantinou$^{*4}$ \quad Georgios Giannarakis$^1$ \\ \textbf{Evagelia Lampiri}$^3$ \quad \textbf{Christos Athanassiou}$^3$ \quad \textbf{Gustau Camps-Valls}$^4$ \\ \textbf{Charalampos Kontoes}$^1$ \quad \textbf{Ioannis Athanasiadis}$^2$\\
$^1$BEYOND Centre, IAASARS, National Observatory of Athens \\\quad $^2$Wageningen University \& Research \\\quad $^3$University of Thessaly \\\quad $^4$Image Processing Laboratory (IPL), Universitat de València \\
\texttt{\{i.tsoumas, giannarakis, kontoes\}@noa.gr}\\
\texttt{\{ilias.tsoumas, ioannis.athanasiadis\}@wur.nl}\\
\texttt{\{elampiri, athanassiou\}@uth.gr}\\
\texttt{\{sitokons, gustau.camps\}@uv.es}
}

\begin{document}

\maketitle
\begin{abstract}
% Pesticides are currently the most widely used solution for pest control in agriculture. However, their extensive use has been shown to contribute to the climate crisis. Integrated pest management (IPM), which incorporates various sustainable practices, is preferred as a climate-smart alternative. Unfortunately, IPM faces low adoption rates as farmers are often skeptical of its effectiveness. In this context, we propose a data analysis framework to enhance IPM. Our framework includes: i) robust pest population predictions across different environments, leveraging invariant and causal learning to address out-of-distribution and data drift issues, ii) interpretable pest presence predictions using glass-box models, which increase trustworthiness and allow for the integration of expert knowledge, iii) counterfactual explanations to provide actionable advice for within-season IPM interventions, iv) heterogeneous treatment effect estimations per field, to assess the suitability of agricultural practices for effective pest prevention and v) causal inference to assess the effectiveness of our advice.
Pesticides serve as a common tool in agricultural pest control but significantly contribute to the climate crisis. To combat this, Integrated Pest Management (IPM) stands as a climate-smart alternative. Despite its potential, IPM faces low adoption rates due to farmers' skepticism about its effectiveness. To address this challenge, we introduce an advanced data analysis framework tailored to enhance IPM adoption. Our framework provides i) robust pest population predictions across diverse environments with invariant and causal learning, ii) interpretable pest presence predictions using transparent models, iii) actionable advice through counterfactual explanations for in-season IPM interventions, iv) field-specific treatment effect estimations, and v) assessments of the effectiveness of our advice using causal inference. By incorporating these features, our framework aims to alleviate skepticism and encourage wider adoption of IPM practices among farmers.
\end{abstract}

\section{Introduction}
% The increasing global population and the changing climate are putting pressure on the agricultural sector, demanding the sustainable production of adequate quantities of nutritious food, feed, and fiber. In this context, we need climate-smart agriculture to optimise crop management with zero waste, enhance resilience, increase production and reduce emissions. In this context, a serious but often underestimated issue is pest management.
% Conventional pest management has been shown to contribute to climate change. The increase in temperature, ultraviolet radiation, and the reduction in relative humidity are expected to increase pest outbreaks and make several pest management practices less effective, including host-plant resistance, bio-pesticides, and synthetic pesticides \cite{sharma2014impact,skendvzic2021impact}. The most common pest management practices rely heavily on pesticides, disregarding the warnings of climate experts. Their extensive use in farming negatively impacts people's health \cite{boedeker2020global} and contributes to the climate crisis through various means. These include: i) greenhouse gas (GHG) emissions from pesticide production, packaging, and transportation \cite{audsley2009estimation}, ii) the impact on soil's ability to sequester carbon \cite{xu2020changing}, iii) the increase in GHG emissions from soil \cite{marty2010volatile,heimpel2013environmental,spokas2003stimulation}, and iv) the contamination of surrounding soil and water sources, leading to loss of biodiversity \cite{sharma2019worldwide}.

Conventional pest management has been shown to contribute to climate change. Raising temperatures, intensifying ultraviolet radiation, and reducing relative humidity, are expected to increase pest outbreaks and undermine the efficacy of pest control methods like host-plant resistance, bio-pesticides, and synthetic pesticides \cite{sharma2014impact,skendvzic2021impact}. Pervasive pesticide use in agriculture, despite climate experts' warnings, adversely affects public health \cite{boedeker2020global} and contributes to the climate crisis. This impact includes: i) greenhouse gas (GHG) emissions from pesticide production, packaging, and transportation \cite{audsley2009estimation}, ii) compromised soil carbon sequestration \cite{xu2020changing}, iii) elevated GHG emissions from soil \cite{marty2010volatile,heimpel2013environmental,spokas2003stimulation}, and iv) contamination of adjacent soil and water ecosystems, resulting in biodiversity loss \cite{sharma2019worldwide}.

Thus, a vicious cycle has been established between pesticides and climate change \cite{vcyclous2022}. In response, the European Commission (EC) has taken action for the reduce of all chemical and high-risk pesticides by 50\% by 2030. Achieving such reductions requires adopting integrated pest management (IPM), which promotes sustainable agriculture and agroecology.
% the European Commission (EC) has taken action through the European Union (EU) Farm to Fork (F2F) strategy. This strategy aims to reduce the overall use of all chemical and high-risk pesticides by 50\% by 2030. Achieving such reductions requires adopting integrated pest management (IPM), which promotes sustainable agriculture and agroecology. IPM encourages using safe alternative methods to protect harvests from pests and diseases. 
IPM consists of 8 principles inspired by the Food and Agriculture Organization (FAO) description. 
% The directive requires all EU member states to ensure its implementation when in parallel Article 55 of Regulation 1107/2009/EC requires that professional pesticide users comply with these principles. 
The authors in \cite{barzman2015eight} condense these principles into prevention and suppression, monitoring, decision-making, non-chemical methods, pesticide selection, reduced pesticide use, anti-resistance strategies, and evaluation.

Data-driven methods have played a crucial role in optimizing pest management decisions. Some employ supervised machine learning (e.g., Random Forests, Neural Networks) with satellite Earth observations (EO) and in-situ data for pest presence prediction \cite{A11,A8}, some incorporating weather data \cite{A7}. Recurrent Neural Networks (RNNs) are used to capture temporal features from weather data, effectively handling unobservable counterfactual outcomes \cite{A4}. Filho et al. extract fine-scale IPM information from meteorological data, insect scouting, remote sensing and machine learning \cite{iost2022does}. Nanushi et al. propose an interpretable machine learning solution for Helicoverpa armigera presence in cotton fields \cite{nanushi2022pest}, enhancing IPM decision-making beyond traditional thresholds. 
% Interpretable predictions boost trust and integrate domain expertise.

% Thus, Artificial Intelligence (AI) and big Earth Observation (EO) data can drive the enhancement of several important IPM’s principles.

% monitoring which consists of , decision-making and evaluation. 
% XAI for IPM ieee paper
% porposal counterfactual explanations + causality AAAI

\section{Proposal}
\begin{figure}[!ht]
\begin{center}
%\framebox[4.0in]{$\;$}
% \fbox{\rule[-.5cm]{0cm}{4cm} \rule[-.5cm]{4cm}{0cm}}
% initial scale 0.55
\includegraphics[scale=0.51]{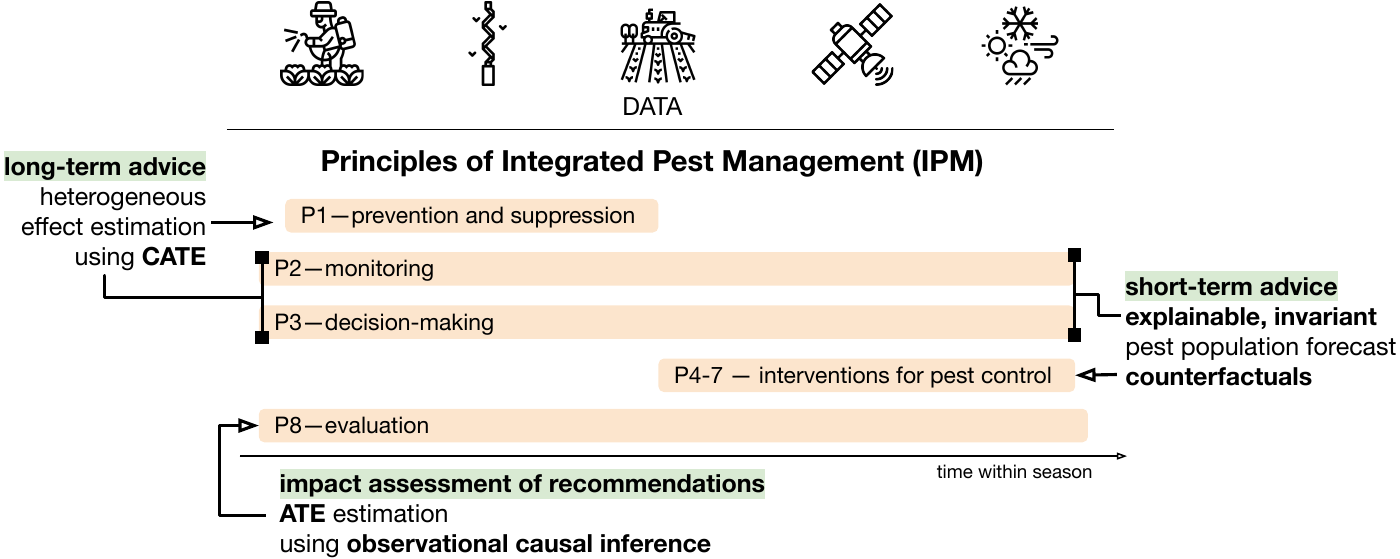}
\end{center}
\caption{Causal and explainable data analysis framework for enhanced IPM}\label{fig:fig1}
\end{figure}

As Barzman et al. point out, threshold-based and "spray/don't spray" advice is not enough \cite{barzman2015eight}. There is a need for a new class of digital tools that take into account the entire set of IPM principles in order to truly enhance decision-making. In this direction, we propose a data analysis framework for IPM based on causality and explainability. It consists of short-term actionable advice for in-season interventions and long-term advice for supporting strategic farm planning (Figure \ref{fig:fig1}). 
% Specifically, the framework will support: 
% i) a 3-5 days forecast of pest population and imminent pest outbreaks, offering global \& local explanations on the predictions and counterfactual explanations in the form of recommended farm interventions,
% ii) the estimation of the heterogeneous effect of IPM practices on pest harmfulness and farm profit.

This way, we will upgrade the \textit{monitoring} and \textit{decision-making} IPM principles leading to actionable advice for direct pest control interventions and assist the selection of practices relevant to other IPM principles, such as \textit{use non-chemical methods} and \textit{reduce pesticide dosage}. Additionally, the proposed framework will better inform farmers with respect to the potential impact of practices that, in turn, will enhance the IPM principle of \textit{prevention and suppression}, e.g., crop rotation, day of sowing, and no-tillage. Furthermore, our framework employs observational causal inference to continuously assess the aforementioned recommendations and thereby satisfy the IPM principle of \textit{evaluation}.

\textbf{Data}: Our strategy hinges on utilizing a variety of data to gain a holistic understanding of historical, current, and future agro-environmental conditions, thereby enhancing our ability to model and comprehend pest dynamics.
% Our proposal relies on diverse data sources to enhance our understanding of agro-environmental conditions and pest dynamics. 
We use EO data on factors like vegetation, soil moisture. Terrain and soil characteristics data are incorporated for long-term area-specific traits. We also utilize weather forecasts and ground measurements, including pest abundance (details in A.1 of Appendix).

\section{Approach \& Methods}

\begin{figure}[!ht]
\begin{floatrow}
\ffigbox{%
\hspace*{-1.7cm}
  \includegraphics[scale=0.195]{
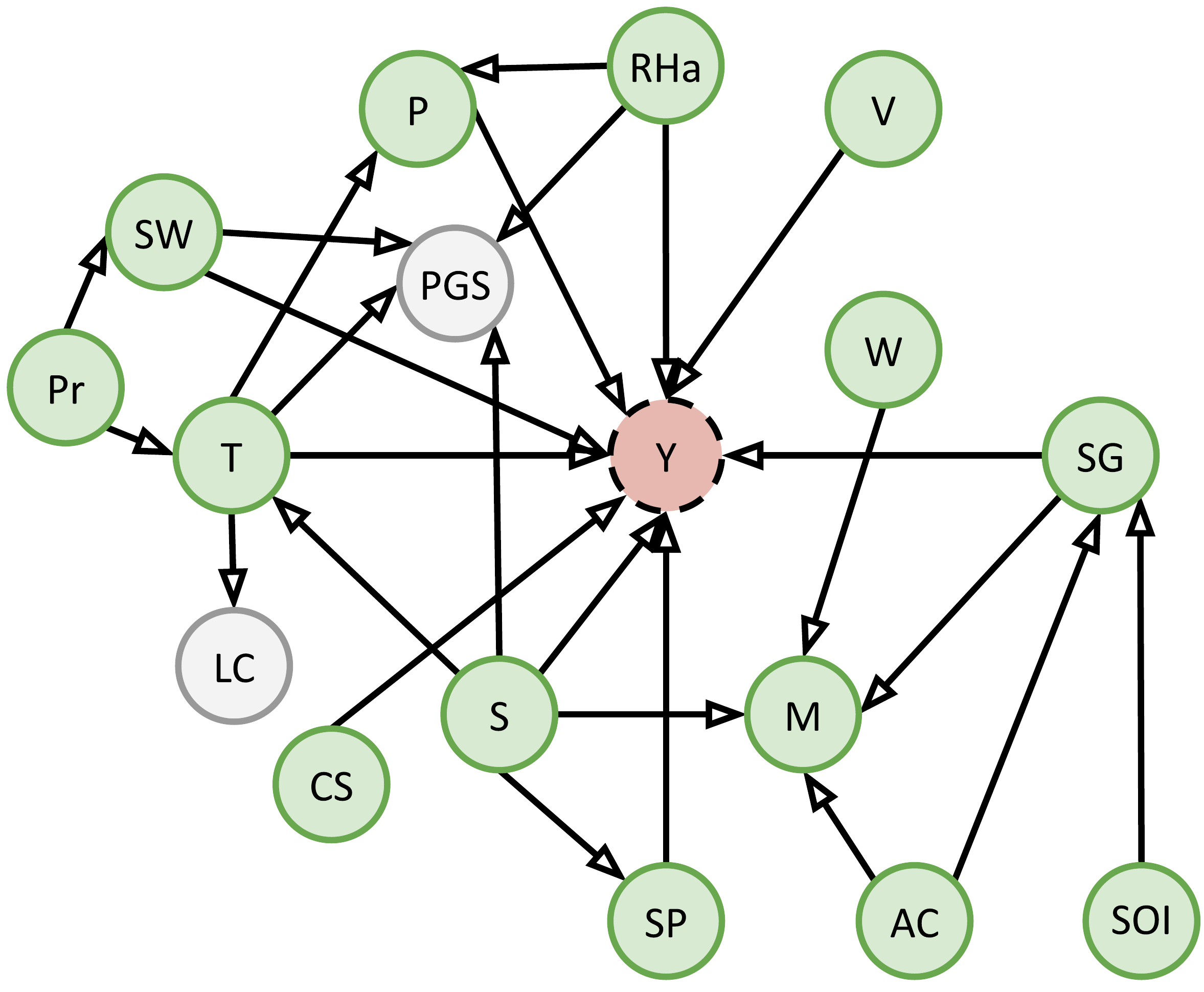}
}{%
    \caption{Causal graph of pest-farm ecosystem. \label{fig:causal-graph}}%
}
\capbtabbox{%
% \begin{table}[!ht]
% \small
% \centering
% \resizebox{\columnwidth}{!}{%
\hspace*{1.0cm}
\renewcommand{\arraystretch}{0.8}
\begin{tabular}{lll}
\toprule
\textbf{Id} & \textbf{Variable Description}             \\ \midrule
T           & Temperature                               \\
SW          & Soil water                                \\
RHa         & Air relative humidity                     \\
SG          & Size of generation                        \\
Pr          & Precipitation                             \\
LC          & Life cycle                                \\
P           & Parasitism                                \\
V           & Variety                                   \\
Sp          & Spraying                                  \\
CS          & Cropping System                          \\
M           & Migration                                 \\
AC          & Adjacent crops                            \\
W           & Wind                                      \\
S           & Season                                    \\
SOI         & South oscillation index                   \\
PGS         & Plant growth stage                        \\
Y           & H. armigera population                    \\ \bottomrule
\end{tabular}%
% \caption{Farm system variable identifier, description and source.}
% \label{tab:variables}
% % }
% \end{table}
}{%
  \caption{Pest-farm ecosystem variables.}%
  \label{fig:variables}
}
\end{floatrow}
\end{figure}

\textbf{Causal Graph for representing domain knowledge.} We constructed a causal graph (Figure \ref{fig:causal-graph}), denoted as $G$, that represents the underlying causal relationships within the pest-farm ecosystem for the H. armigera case. The graph $G$ comprises vertices $V$, which represent the variables in the system, and directed edges $E$, which symbolize the cause-and-effect relationships between these variables (details about graph building in A.2 of Appendix). Besides helping us articulate domain knowledge, the causal graph $G$ will benefit the downstream technical analyses in various ways.
$G$ serves as an amalgamation of domain knowledge and a foundational schema that can be leveraged variably depending on the methodological requirements of the analytical techniques in play. 

\textbf{Invariant \& Causal Learning for Robust Pest Prediction.} Our goal is to predict near-future pest populations ($Y_{t+1}$) using EO and environmental data ($X_{t}$) and weather forecasts ($W_{t+1}$) by learning the function $y_{t+1} = f(x_{t}, w_{t+1})$. Conventional machine learning methods \cite{A11,A8,A7,A4} struggle with non-i.i.d. data, hindering generalization and adaptation. We turn to causal learning \cite{scholkopf2022statistical}, grounded in independent causal mechanisms that remain stable despite environmental changes. To achieve this, we integrate invariant learning with causality, categorizing data into environments $E$ (e.g., agroclimatic zones). While $E$ influences features ($x_{t},w_{t+1}$), it does not directly affect the target ($Y_{t}$). Invariant Causal Prediction (ICP) \cite{heinze2018invariant}, DAGs, and Invariant Risk Minimization (IRM) \cite{arjovsky2019invariant} help select causal features, identify potential relationships, and capture latent causal structures.

\textbf{Explainability \& Counterfactual Reasoning for Short-term Advice.} We define the problem as a binary classification of pest presence or absence, at the next time step, using EO data ($X_{t}$) and weather forecasts ($W_{t+1}$). We employ Explainable Boosting Machines (EBM) \cite{nori2019interpretml} to enhance predictions with explanations at global and local levels. EBM's additive model allows visualization of feature contributions, enhancing trust. To bolster trust, we propose generating counterfactual examples as recommended interventions. We follow the setup of \cite{mothilal2020explaining}, searching for minimal feature perturbations in $(x_{t},w_{t+1})$ that alter predictions using the same model $f$. These counterfactual examples represent proposed actions for real farm systems, ensuring practicality and feasibility \cite{wachter2017counterfactual,mothilal2020explaining}.

% \textbf{Heterogeneous Treatment Effects for long-term IPM advice. }
% We enable long-term advice for\textit{ prevention and suppression} of harmful pest populations via assessing the heterogeneous effects of pertinent practices (e.g., crop rotation, balanced fertilization, stale seedbed technique, sowing dates) on combined pest harmfulness and yield indices. Different fields, characterized by other agro-environmental conditions, may respond differently to the same practice. Hence, we estimate the conditional average treatment effect (CATE), following the formulation of \cite{giannarakis2022towards}, where the authors assessed the heterogeneous impact of crop rotation and landscape crop diversity on net primary productivity and soil organic carbon \cite{giannarakis2022personalizing}. Using the potential outcomes framework \cite{rubin2005causal}, let $Y(T)$ denote the value of a random variable Y (i.e., pest harmfulness and yield) if we were to treat a unit with a treatment T $\in \{0, 1\}$. CATE is estimated as the difference of potential outcomes $\mathbb{E}[Y(T=1)-Y(T=0)|X]$, controlling for $X$, consisting of the field characteristics (see data subsection) that drive heterogeneity.

\textbf{Heterogeneous Treatment Effects for Long-term Advice.} We provide long-term advice for pest prevention and suppression by assessing how practices (e.g., crop rotation, balanced fertilization, sowing dates) affect pest harmfulness and yield indices. Different agro-environments may yield varying responses to the same practice. We estimate the conditional average treatment effect (CATE) \cite{giannarakis2022towards} using the potential outcomes framework \cite{rubin2005causal}. CATE quantifies the difference in potential outcomes ($\mathbb{E}[Y(T=1)-Y(T=0)|X]$), controlling for field characteristics that drive heterogeneity.

\textbf{Causal Inference for Evaluating Advice Effectiveness.} We employ causal inference techniques to assess the effectiveness of our pest control recommendations, building on a recent approach introduced in the context of cotton farming \cite{tsoumas2022evaluating}. Adapting this method to pest control interventions, we turn to difference-in-differences \cite{abadie2005semiparametric}. Our aim is to quantify the average treatment effect of adhering to our framework's recommendations (\textit{treated units}) compared to those who did not (\textit{control units}). Historical intervention data, annotated as recommended or not, will be used for the evaluation. Causal inference will be conducted on a per-environment basis, ensuring similarity between treatment and control groups, following the parallel trends assumption \cite{lechner2011estimation}. Depending on data volume and time series length, other methods like synthetic control or panel data may also be considered.

\section{Conclusions}
% Breaking the harmful cycle between pesticides and climate change is essential. In this direction, IPM aims to successfully control pests while minimising the adverse effects of conventional pest management on human health and the environment. We propose an AI-driven framework for IPM that provides short- and long-term advice, promoting sustainable practices and timely control methods. Our approach prioritizes transparency and fairness, ensuring effective pest control in agriculture. Additionally, we employ observational causal inference to evaluate the framework's effectiveness, enhancing trust and transparency.
Breaking the harmful cycle between pesticides and climate change is essential. In this direction, IPM aims to successfully control pests while minimizing the adverse effects of conventional pest management on human health and the environment. We propose an AI-driven framework for IPM that provides short- and long-term advice, promoting sustainable practices and timely control methods. Additionally, we employ observational causal inference to evaluate the framework's effectiveness. Finally, our approach ensures effective pest control and enhances trust and transparency.

\begin{ack}
We express our gratitude to Corteva Agriscience Hellas, particularly to Dr. George Zanakis, the Marketing \& Development Manager, for their invaluable support, trust, and provision of data. This research was primarily funded by the "Financing of Charalambos Kontoe's Research Activities\_code 8003" under Special Reseacrh Account of National Observatory of Athens. I. N. Athanasiadis work has been partially supported by the European Union Horizon 2020 Research and Innovation program (Project Code: 101070496, Smart Droplets). Vasileios Sitokonstantinou and Gustau Camps-Valls work has been supported by the GVA PROMETEO project "Artificial Intelligence for complex systems: Brain, Earth, Climate, Society" agreement CIPROM/2021/56.

\end{ack}

\bibliographystyle{splncs04}
\bibliography{refs.bib}

\appendix{
\section{Supplementary Material}
\subsection{Data}
% eo data, meteo data, corteva data
Our approach relies on diverse data sources as a key leverage to capture a comprehensive picture of the past, present, and future agro-environmental conditions. In turn, this will enable us to improve the modeling and comprehension of pest dynamics.

\textbf{Earth Observations}: We leverage biophysical and biochemical properties such as Leaf Area Index (LAI), Normalized Difference Vegetation Index (NDVI), chlorophyll content, as well as data on evapotranspiration and soil moisture. These factors play a crucial role in monitoring pest population dynamics. The data is derived from the Sentinel-1/2 and Terra/Aqua (MODIS) satellite missions that provide open access to optical multi-spectral and Synthetic Aperture Radar (SAR) images.

\textbf{Terrain \& soil characteristics}: We incorporate data from open-access digital elevation models, as well as information on topsoil physical properties and soil organic carbon content \cite{de2015map,ballabio2016mapping}. This allows us to include fixed or long-term characteristics specific to the area of interest.

\textbf{Numerical weather predictions (NWP) and reanalysis environmental datasets}: We utilize a custom configuration of WRF-ARW \cite{skamarock2019description} at a spatial resolution of 2 km. Hourly predictions are made, and for each trap location, we obtain daily values for air (2 m) and soil temperature (0 m), relative humidity (RH), accumulated precipitation (AP), dew point (DP), and wind speed (WS). These parameters have been widely used in related work and are extremely valuable for learning from past (reanalysis) and future (NWP) pest states.

\textbf{In-field measurements}: In-field measurements involve ground observations of pest abundance using pheromone traps specifically designed for monitoring the cotton bollworm, known by the scientific name Helicoverpa armigera (H. armigera). These traps contain the active ingredients Z-11-hexadecen-1-al and Z-9-hexadecenal. The traps are used from the beginning of the first generation until the end of the season, with regular replacement every 4 to 6 weeks. The company Corteva Agriscience Hellas has established a dense (in time and space) trap network (Figure \ref{fig:fig2}) that covers almost all areas in the Greek mainland where cotton is cultivated. The traps are strategically positioned at suitable distances from each other to prevent interference and ensure accurate data collection. An agronomist examines the traps and counts the trapped insects at regular intervals every 3-5 days. Corteva Agriscience Hellas provides us with historical data consisting of 398 trap sequences and 8202 unique data points since 2019 (Table \ref{fig:data}). They also provide auxiliary data on pesticide application, potential crop damage from pests, the severity of the damage, trap replacements, and scouter comments.

\begin{table}[!h]
\centering
\resizebox{11cm}{!}{%
\begin{tabular}{@{}cccccccc@{}}
\toprule
\textbf{Year} & \textbf{Traps} & \textbf{Measurements} & \textbf{Mean} & \textbf{std} & \textbf{Sprays} & \textbf{\begin{tabular}[c]{@{}c@{}}Sprayed \\ fields \%\end{tabular}} \\ \midrule
2022          & 126            & 2507                 & 19.73         & 4.22         & 30              & 18.25                                                                                                                                     \\
2021          & 109            & 2245                 & 20.30         & 1.79         & 17              & 11.01                                                                                                                                     \\
2020          & 81             & 1693                 & 20.54         & 4.77         & 12              & 8.64                                                                                                                                     \\
2019          & 82             & 1757                 & 21.29         & 6.43         & 21              & 21.95                                                                                                                                    \\ \bottomrule
\end{tabular}
}

% \caption{Trap data summary. Agroclimatic zones column is referred in the different zones that dataset's traps are located. Zones identification based on \cite{ceglar2019observed}} 
\caption{Summary of Trap Data.}
\label{fig:data}
\end{table}

% \usepackage{booktabs}
% \usepackage{graphicx}
% \begin{table}[]
% \centering
% \resizebox{\textwidth}{!}{%
% \begin{tabular}{@{}cccccccc@{}}
% \toprule
% \textbf{year} & \textbf{traps} & \textbf{measurments} & \textbf{mean} & \textbf{std} & \textbf{sprays} & \textbf{\begin{tabular}[c]{@{}c@{}}sprayed \\ fields \%\end{tabular}} & \textbf{\begin{tabular}[c]{@{}c@{}}agroclimatic\\ zones\end{tabular}} \\ \midrule
% 2022 & 126 & 2507 & 19.73 & 4.22 & 30 & 18.25 & 3 \\
% 2021 & 109 & 2245 & 20.30 & 1.79 & 17 & 11.01 & 3 \\
% 2020 & 81 & 1693 & 20.54 & 4.77 & 12 & 8.64 & 3 \\
% 2019 & 82 & 1757 & 21.29 & 6.43 & 21 & 21.95 & 3 \\ \bottomrule
% \end{tabular}%
% }
% \caption{Trap data summary. \label{fig:data}}
% \end{table}

% \begin{figure}[!ht]
% \begin{center}
% %\framebox[4.0in]{$\;$}
% % \fbox{\rule[-.5cm]{0cm}{4cm} \rule[-.5cm]{4cm}{0cm}}
% \includegraphics[scale=0.55]{draft_traps_images.png}
% \end{center}
% \caption{Traps in the Greek mainland for the period 2019-2022}\label{fig:fig2}
% \end{figure}

\begin{figure}[!ht]
\begin{center}
%\framebox[4.0in]{$\;$}
% \fbox{\rule[-.5cm]{0cm}{4cm} \rule[-.5cm]{4cm}{0cm}}
\includegraphics[scale=0.65]{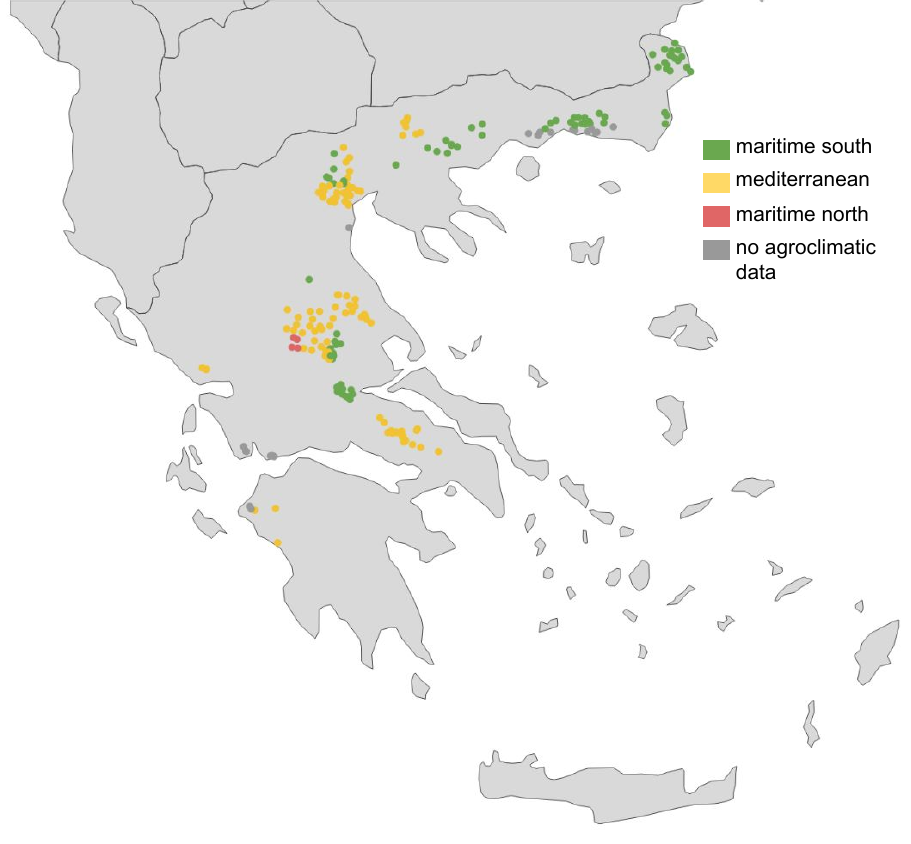}
\end{center}
\caption{Traps distribution in the Greek mainland for the period 2019-2022. Colors indicate the different agroclimatic zones in which traps from the dataset belong. These zones have been identified based on the study conducted by Ceglar et al. \cite{ceglar2019observed}.}\label{fig:fig2}
\end{figure}

\subsection{Domain Knowledge and Graph Building.} 
In the current case about the pest-farm ecosystem of H. armigera, various biotic and abiotic factors (Table \ref{fig:variables}) can influence the population dynamics $Y$ of H. armigera \cite{sharma2012monitoring}. Temperature $T$ plays a crucial role, affecting the growth, development, fecundity, and survival of the insect \cite{howe1967temperature}. The size $SG$ of the first generation is related to the size of the second generation, and the Southern Oscillation Index $SOI$ has a significant correlation with the size of the first spring generation \cite{maelzer1999analysis,maelzer2000long}. Additionally, the life cycle $LC$ of H. armigera is temperature-dependent, with completion occurring between 17.5°C and 32.5°C \cite{mironidis2014development}. Depending on the season, the life cycle can be completed within 4–6 weeks in summer, increasing to 8–12 weeks in autumn \cite{ali2009some}. The presence of parasitoids and natural enemies in cotton cultivation, is a crucial component of many IPM programs, including the control of H. armigera \cite{pereira2019potential}. Many egg parasitoids of the different families are known for their high parasitism $P$ rates and their effectiveness in reducing the population of H. armigera \cite{noor2015evaluation}. Nevertheless, parasitism rates are influenced by temperature and relative humidity \cite{kalyebi2005parasitism,noor2015evaluation}. Moreover, the efficacy of spray application $Sp$ also impacts population dynamics \cite{wardhaugh1980incidence}.

Other environmental factors come into play as well. Precipitation $Pr$ affects the population size, with heavy precipitation leading to a decrease in the population \cite{ge2003life}. It also increases air relative humidity $RHa$ and soil water content $SW$, that in their turn affect the emergence rate of H. armigera \cite{fajun2003effects}. The presence of fruiting organs during the plant growth stage $PGS$ is important for population dynamics, as it serves as the oviposition site for females \cite{fitt1989ecology}. Crop variety $V$, such as transgenic Bt cotton, can suppress the second generation of H. armigera, while both different cropping systems $CS$ and adjacent crops $AC$ can influence the population structure \cite{wardhaugh1980incidence,gao2010active,lu2013towards}. Finally, wind $W$ and wind direction play a significant role in the seasonal migration $M$ of H. armigera, impacting the distance covered during migration \cite{torres2002pyrethroid,feng2005high}. These various factors collectively shape the population dynamics of H. armigera in a complex and interconnected manner.

}

\end{document}